\pdfoutput=1

\documentclass[11pt]{article}

\usepackage[]{ACL2023}

\usepackage{times}
\usepackage{latexsym}
\usepackage[most]{tcolorbox}
\usepackage{amsmath}
\usepackage{graphicx}
\usepackage{multirow}
\usepackage{booktabs}
\usepackage{colortbl}
\usepackage{xcolor}
\usepackage{tabularx}
\usepackage{subcaption}
\usepackage{algorithm}
\usepackage{algorithmic}
\usepackage{amsmath}
\usepackage{amsfonts}
\usepackage{amssymb}
\usepackage{CJKutf8}

\usepackage{tablefootnote}
\usepackage[T1]{fontenc}

\usepackage[utf8]{inputenc}
\usepackage{amsmath}
\usepackage{microtype}
\usepackage{enumitem}
\usepackage{inconsolata}

\usepackage[colorinlistoftodos,prependcaption,textsize=small]{todonotes}
\presetkeys{todonotes}{inline}{}

\usepackage[normalem]{ulem}
\usepackage{seqsplit}

\title{CLEV: LLM-Based Evaluation Through Lightweight Efficient Voting for Free-Form Question-Answering}
\author{
  Sher Badshah\textsuperscript{1},
  Moamen Moustafa\textsuperscript{2},
  Hassan Sajjad\textsuperscript{1} \\
  \textsuperscript{1}Dalhousie University \\
  \textsuperscript{2}Alamein International University \\
  \texttt{\{sh545346, hsajjad\}@dal.ca} \quad
  \texttt{mmoustafa@aiu.edu.eg}
}
\begin{document}

\maketitle
\begin{abstract}
Evaluating free-form Question-Answering (QA) remains a challenge due to its diverse and open-ended nature. Traditional automatic metrics fail to capture semantic equivalence or accommodate the variability of open-ended responses. Leveraging Large Language Models (LLMs) as evaluators offers a promising alternative due to their strong language understanding and instruction-following capabilities. We propose the Consensus via Lightweight Efficient Voting (CLEV), which employs two primary LLMs as judges and engages a third judge only in cases of disagreement. This approach prioritizes evaluation reliability while reducing unnecessary computational demands. Through experiments, including human evaluation, we demonstrate CLEV’s ability to provide consistent, scalable, and resource-efficient assessments, establishing it as a robust framework for evaluating LLMs on free-form QA.
\end{abstract}

\section{Introduction}
One of the practical limitations in evaluating free‑form Question‑Answering (QA) is the lexical–semantic mismatch between Large Language Model (LLM) outputs and the reference answers. For the query \textit{``Who wrote 1984?''}, a dataset may list simply \textit{``George Orwell''}, while a helpful model replies: \textit{``It was penned by the British author Eric Arthur Blair.''} Although both refer to the same person, there is no surface token overlap, so lexical-matching and even embedding-based metrics assign it an unduly low score. For instance, Exact Match (EM) requires strict lexical alignment (e.g., failing to equate ``nuclear weapon'' and ``atomic bomb'') and ignores semantic equivalence~\citep{doostmohammadi2024reliable}.

A reference-aware LLM-as-a-judge can instead reason over meaning, recognize that the candidate entails the gold fact, and deliver a verdict, thereby overcoming this lexical–semantic gap~\citep{10.5555/3666122.3668142, verga2024replacing}. Existing studies using LLM as judges primarily focus on subjective pairwise comparison~\citep{wang2023large, vu2024autoraterstaminglarge} and single-answer scoring~\citep{chianglee2023large, hu2024llmbased, liuetal2023g}. However, objective evaluation using LLM judges, particularly for free-form QA, has received limited attention. Furthermore, LLM-based judging also lies on a cost–quality spectrum. Querying a single judge is efficient but is less reliable due to the known limitations, such as prompt sensitivity, inconsistency, and bias \citep{ye2024justiceprejudicequantifyingbiases}, as no individual model captures the full diversity of reasoning styles, long-tail knowledge, and user values~\citep{feng2025llmdroolsmultillmcollaboration}. To improve robustness, some studies employ fixed ensembles of three or more LLM judges and aggregate their decisions via majority voting~\citep{badshah-sajjad-2025-reference, verga2024replacing}. While this improves reliability, it increases computational cost and latency~\citep{2024quantifyingcapabilitiesllmsscale}, limiting scalability for large-scale evaluation~\citep{jung2024trustescalatellmjudges, adlakha2024evaluating}.

To address these trade-offs, we propose the Consensus via Lightweight Efficient Voting (CLEV) that balances the reliability and efficiency of using LLMs as judges for free-form QA. CLEV employs two primary judges for initial assessments and invokes a third judge only when disagreements occur. By minimizing redundant calls in the fixed three-judge majority vote setting, CLEV reduces computational overhead by roughly 80 to 95\% (varying by task) while achieving substantial to perfect agreement. Our key contributions include: 1) establishing a principled LLM-as-a-judge evaluation setup for free-form QA that moves beyond token-level string matching metrics, 2) proposing CLEV, a lightweight consensus-based evaluation method that preserves the reliability of multi-judge voting while significantly reducing computational cost, 3) empirically validating CLEV across diverse QA datasets and multiple state-of-the-art LLMs, and 4) systematic analysis of failure cases.

\begin{figure*}[ht] 
\centering
\includegraphics[width=0.99\textwidth]{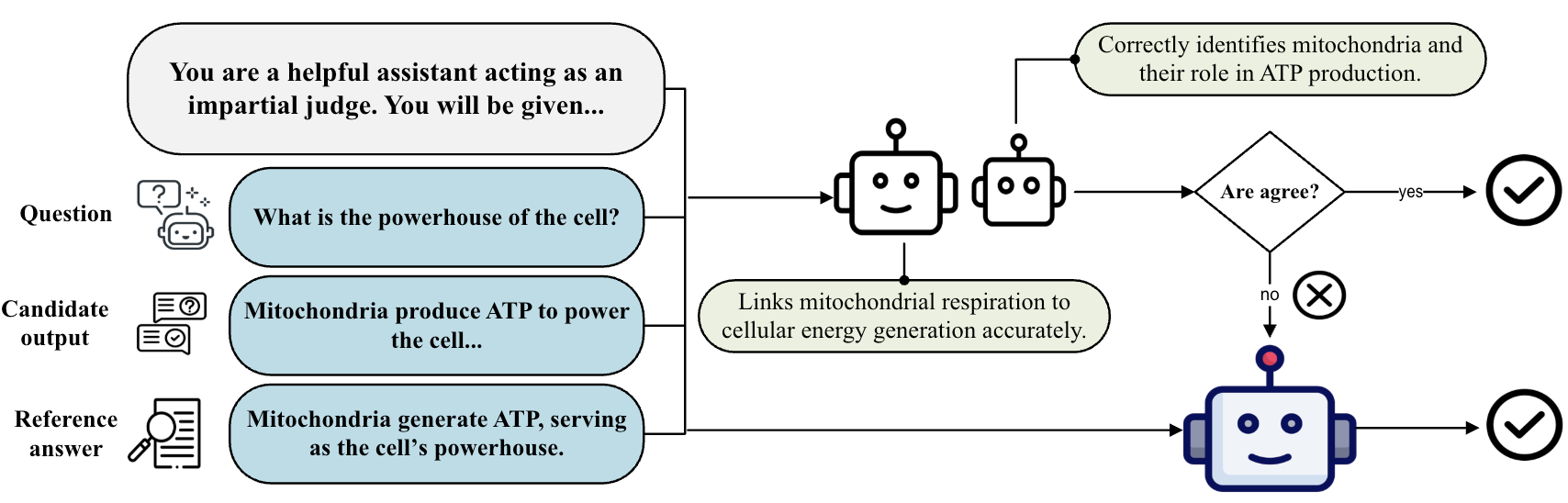}
\caption{Our proposed Consensus via Lightweight Efficient Voting.}
\label{fig:method}
\end{figure*}

\section{Methodology}
This section outlines the core components of our evaluation framework. Given a question and a candidate model’s answer, the evaluation task is to determine whether the answer is factually correct with respect to a reference. Because free-form QA requires an objective correctness decision rather than scalar scoring, we prompt LLM judges to return a binary verdict along with a brief rationale explaining their judgment. The verdict serves as the evaluation signal, while the rationale provides interpretability and supports later analysis.

\subsection{Consensus via Lightweight Efficient Voting (CLEV)}\label{sec:CLEV_method}
In traditional human evaluation settings, when two annotators disagree on a judgment, a third expert is often called upon to resolve the dispute. Drawing inspiration from this efficient practice, we propose CLEV. As illustrated in Figure~\ref{fig:method}, rather than immediately employing three LLMs, CLEV adopts an efficient approach by beginning with two models as primary judges. When these judges reach a consensus, no further evaluation is needed. Only in cases of disagreement, the third LLM is engaged, whose decision then creates a majority verdict.

\subsection{Judges inclusion and exclusion criteria}\label{judges_inc_exc}
\begin{figure}[t]
\centering
\includegraphics[width=\linewidth,keepaspectratio]{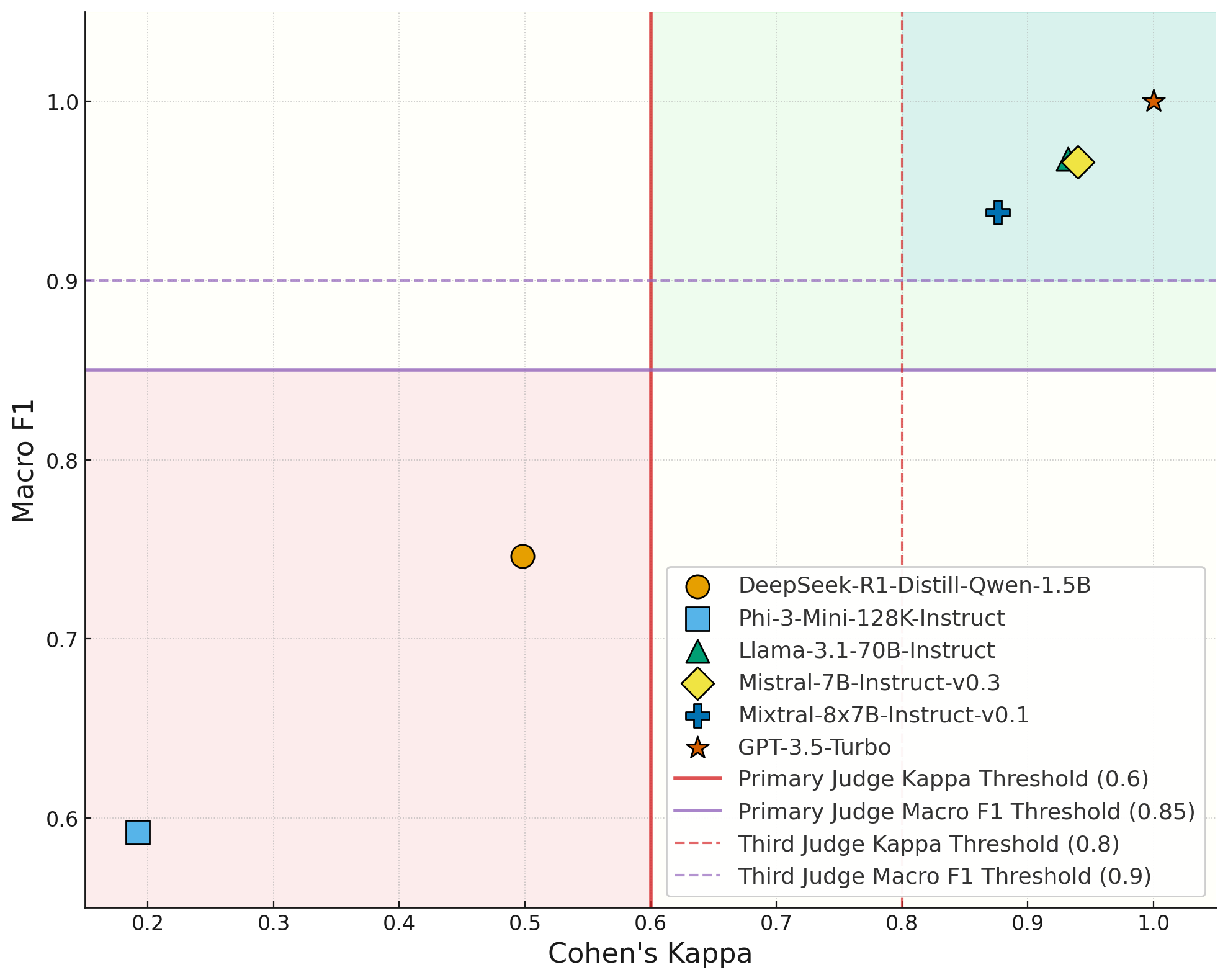}
\caption{Judges selection based on the defined criteria in Section~\ref{judges_inc_exc}.}
\label{fig:incl_excl}
\end{figure}

To systematically select suitable judges for CLEV, we evaluate various LLMs (see Figure~\ref{fig:incl_excl}) using 100 random instances from HotpotQA. For each model, we compare binary verdicts against human annotations and compute Cohen’s Kappa ($\kappa$) and Macro F1. We interpret $\kappa$ following the commonly used guideline where values between $0.61$–$0.80$ indicate \emph{substantial agreement}, and values above $0.80$ indicate \emph{near-perfect agreement}~\citep{mchugh2012interrater}. However, since $\kappa$ is known to be sensitive to class imbalance~\citep{cicchetti1990high}, we together consider Macro F1 to ensure balanced evaluation across both classes:

{\footnotesize
\[
\text{status} =
\begin{cases}
  \text{primary judges}, & \kappa \geq 0.6 \ \text{and} \ \text{F1} \geq 0.85, \\[4pt]
  \text{third judge}, & \kappa \geq 0.8 \ \text{and} \ \text{F1} \geq 0.9, \\[4pt]
  \text{excluded}, & \text{otherwise}.
\end{cases}
\]
}

\section{Experiments}
We utilize the following settings to evaluate CLEV:
\paragraph{Models} Our candidate models includes Llama-3.1 70B~\citep{MetaLlama3_2024}, GPT-3.5-turbo~\citep{NEURIPS2020_1457c0d6}, Mistral 7B~\citep{jiang2023mistral}, and Mixtral 8x7B~\citep{jiang2024mixtralexperts}. For judges, Mistral‑7B met our inclusion criteria ($\kappa \geq 0.6$, $\mathrm{F1} \geq 0.85$) while offering low cost and fast inference. Llama‑3.1‑70B, though more expensive, provided substantial agreement. We therefore use both as primary judges to capture a range of capabilities~\citep{feng2025llmdroolsmultillmcollaboration, liangetal2024encouraging, sun2024skillaggregation}. GPT‑3.5‑turbo serves as the third judge because it exceeds the required thresholds ($\kappa \geq 0.8$, $\mathrm{F1} \geq 0.9$). All models are run with temperature 0 to ensure reproducibility, as higher temperatures degrade evaluator reliability~\citep{hada2024llmbasedevaluators}.

\paragraph{Datasets} We utilize four datasets: AmbigQA~\citep{minetal2020ambigqa}, HotpotQA~\citep{yang2018hotpotqa}, Natural Questions~\citep{kwiatkowskietal2019}, TriviaQA~\citep{joshi2017triviaqa}. We only utilize FreshQA to evaluate the judge's ability to detect outdated information. See Appendix~\ref{app:datasets} for details.

\paragraph{Prompts} We design minimal zero-shot role-playing prompts~\citep{kong2024groleplay} for both candidate and judge LLMs. The judge LLM is instructed to return a binary verdict with a brief explanation, which reduces subjectivity and simplifies automatic evaluation (see Appendix~\ref{app:prompting}).

\paragraph{Baselines} We compare individual LLM judges and CLEV against EM, BERTScore~\citep{ZhangKWWA20} computed with microsoft/deberta-xlarge-mnli and thresholded at \(\tau=0.5\) to yield binary decisions and majority voting, which always invokes a fixed setup of LLM judges and returns the model verdict (see Appendix~\ref{app:baselines}).

\paragraph{Human evaluation}
We invite three volunteer graduate students from our lab to act as annotators. We provided the questions, reference answers, and candidate LLM responses, without information about model identity, to avoid bias. Each response is scored on a binary scale based on correctness and relevance (see Appendix~\ref{app:human_evaluation}).

\paragraph{Evaluation metrics}
We compute Fleiss’ Kappa (\(\kappa\))~\citep{fleiss1973equivalence} and percent agreement to measure inter-rater reliability among human annotators. To compare evaluators with humans, we calculate Cohen’s Kappa~\citep{mchugh2012interrater} against the human majority on an instance level. Because of class imbalance, kappa can appear low despite high agreement~\citep{cicchetti1990high}. Therefore, we also frame the task as binary classification and report Macro-F1 scores.

\section{Results}
In this section, we briefly report the results and refer the readers to Appendix~\ref{app:additional_results} for detailed results.

\begin{table*}[t]
\centering
\scriptsize
\setlength{\tabcolsep}{4pt}
\resizebox{\textwidth}{!}{
\begin{tabular}{llcccccccc}
\toprule
 & & \multicolumn{7}{c}{\textbf{Evaluators (Cohen's $\kappa$ [Macro-F1] vs Human Majority)}} \\
\cmidrule(lr){3-9}
\textbf{Candid.} & \textbf{Tasks} & \textbf{EM} & \textbf{BERTScore} &
\textbf{Llama} & \textbf{GPT} & \textbf{Mistral} & \textbf{MV} & \textbf{CLEV} & \textbf{Disagr. (\%)} \\
\midrule
\multirow{4}{*}{Llama}
& AmbigQA  & 0.52 [0.74] & 0.28 [0.64] & 0.89 [0.94] & 0.84 [0.92] & 0.86 [0.93] & 0.91 [0.96] & \cellcolor{green!15}0.91 [0.96] & 10.0 \\
& HotpotQA & 0.58 [0.78] & 0.50 [0.75] & 0.88 [0.94] & 0.90 [0.95] & 0.83 [0.92] & 0.95 [0.98] & \cellcolor{green!15}0.95 [0.98] & 13.0 \\
& NQ-Open  & 0.38 [0.65] & 0.44 [0.72] & 0.83 [0.92] & 0.79 [0.90] & 0.74 [0.87] & 0.93 [0.96] & \cellcolor{green!15}0.92 [0.96] & 18.0 \\
& TriviaQA & 0.28 [0.61] & 0.56 [0.78] & 0.55 [0.77] & 0.44 [0.72] & 0.30 [0.64] & 0.68 [0.84] & \cellcolor{green!8}0.68 [0.84] & 17.0 \\
\midrule
\multirow{4}{*}{GPT}
& AmbigQA  & 0.56 [0.79] & 0.25 [0.62] & 0.94 [0.97] & 0.90 [0.95] & 0.85 [0.93] & 0.97 [0.98] & \cellcolor{green!15}0.96 [0.98] & 7.0 \\
& HotpotQA & 0.60 [0.79] & 0.30 [0.62] & 0.95 [0.98] & 0.97 [0.99] & 0.93 [0.97] & 0.99 [0.99] & \cellcolor{green!15}0.98 [0.99] & 5.7 \\
& NQ-Open  & 0.45 [0.70] & 0.22 [0.61] & 0.88 [0.94] & 0.82 [0.91] & 0.83 [0.91] & 0.96 [0.98] & \cellcolor{green!15}0.95 [0.98] & 13.0 \\
& TriviaQA & 0.34 [0.65] & 0.36 [0.68] & 0.65 [0.82] & 0.40 [0.70] & 0.47 [0.73] & 0.81 [0.90] & \cellcolor{green!8}0.77 [0.89] & 15.7 \\
\midrule
\multirow{4}{*}{Mixtral}
& AmbigQA  & 0.55 [0.76] & 0.34 [0.67] & 0.90 [0.95] & 0.78 [0.89] & 0.89 [0.94] & 0.98 [0.99] & \cellcolor{green!15}0.95 [0.98] & 9.0 \\
& HotpotQA & 0.55 [0.76] & 0.35 [0.66] & 0.94 [0.97] & 0.93 [0.97] & 0.94 [0.97] & 0.97 [0.99] & \cellcolor{green!15}0.97 [0.99] & 4.7 \\
& NQ-Open  & 0.37 [0.65] & 0.30 [0.65] & 0.88 [0.94] & 0.73 [0.86] & 0.82 [0.91] & 0.97 [0.98] & \cellcolor{green!15}0.91 [0.96] & 13.0 \\
& TriviaQA & 0.32 [0.63] & 0.39 [0.70] & 0.63 [0.81] & 0.61 [0.80] & 0.44 [0.72] & 0.90 [0.95] & \cellcolor{green!8}0.76 [0.88] & 17.0 \\
\midrule
\multirow{4}{*}{Mistral}
& AmbigQA  & 0.60 [0.79] & 0.25 [0.62] & 0.89 [0.95] & 0.89 [0.95] & 0.86 [0.93] & 0.95 [0.98] & \cellcolor{green!15}0.95 [0.98] & 11.7 \\
& HotpotQA & 0.61 [0.80] & 0.38 [0.67] & 0.94 [0.97] & 0.90 [0.95] & 0.94 [0.97] & 0.96 [0.98] & \cellcolor{green!15}0.95 [0.98] & 6.0 \\
& NQ-Open  & 0.48 [0.73] & 0.29 [0.64] & 0.85 [0.93] & 0.84 [0.92] & 0.84 [0.92] & 0.95 [0.98] & \cellcolor{green!15}0.95 [0.98] & 14.7 \\
& TriviaQA & 0.47 [0.72] & 0.24 [0.61] & 0.76 [0.88] & 0.73 [0.86] & 0.47 [0.74] & 0.87 [0.94] & \cellcolor{green!8}0.85 [0.93] & 20.3 \\
\bottomrule
\end{tabular}
}
\caption{Cohen's Kappa and [Macro-F1] scores showing the agreement of individual evaluators, Majority Vote (MV), and CLEV with human majority across models and tasks. \textbf{Disagr. (\%)} indicates the percentage of disagreements in the majority voting (Llama+GPT+Mistral), i.e., when the third judge is called in CLEV.}
\label{tab:combined_kappa_macro_f1_final}
\end{table*}

\paragraph{Correlation with human evaluation} As depicted by consistently high Cohen's kappa in Table~\ref{tab:combined_kappa_macro_f1_final}, CLEV maintains strong agreement with human evaluation. This represents an improvement over individual model performance, where individual judges generally showed varying levels of agreement with human evaluation. 

\noindent\textbf{LLM-based evaluators demonstrate strong abilities in recognizing semantic variations} while maintaining the core meaning, especially when assessing responses that use different terminology or structural approaches to convey the same information. For instance, evaluators correctly identified that \textit{``Salma Hayek''} and \textit{``Salma Hayek Pinault''} refer to the same individual, acknowledging the semantic equivalence despite differences in phrasing. Similarly, when assessing responses that use different terms for the same entity, such as recognizing \textit{``Nick Fury, Agent of S.H.I.E.L.D.''} as part of the broader \textit{``Marvel''} universe, the evaluators effectively maintain the core meaning and contextual relevance. Their explanations show systematic assessment patterns that combine multiple evaluation criteria (e.g., factual accuracy and contextual relevance).

\noindent\textbf{CLEV reducing third-judge calls by 80–95\%} while maintaining the performance across tasks. As depicted in Table~\ref{tab:combined_kappa_macro_f1_final}, even on TriviaQA, where disagreement is high, CLEV scores remain in substantial to perfect agreement. 

\noindent\textbf{EM underestimates and BERTScore overestimates.} Interestingly, EM typically accomplishes better correlation with human evaluation on the instance-level in Table~\ref{tab:combined_kappa_macro_f1_final} than neural-based BERTScore. EM's strict nature leads to lower overall performance, but its high precision ensures that when it identifies a match, it strongly aligns with human annotations.  In contrast, BERTScore takes a more lenient approach to semantic matching. As a result, it shows more false positives, consequently reducing instance-level agreement with human judgments.

\subsection{Impact of Selective Third Judge}
A core benefit of CLEV is that it avoids unnecessary third-judge evaluations by only invoking a judge when the two primary judges disagree. Table~\ref{tab:clev_agg_escalation} summarizes the frequency of such escalation events across five datasets and five candidate LLMs. Across all 7{,}500 evaluated instances (300 examples $\times$ 5 models $\times$ 5 tasks), only 17.6\% required a third judge. This reduction is substantial when compared to the fixed three-judge majority voting setup, which always incurs 100\% of third-judge calls. In other words, CLEV reduces evaluation overhead by approximately 80--95\% while preserving agreement levels that match majority vote outcomes (see Table~\ref{tab:combined_kappa_macro_f1_final}). Tasks with stable factual grounding (e.g., HotpotQA) trigger few escalations, whereas more temporally dynamic datasets like FreshQA show higher disagreement rates due to shifts in real-world facts. 

\begin{table}[t]
\centering
\footnotesize
\setlength{\tabcolsep}{8pt}
\begin{tabular}{lcc}
\toprule
\textbf{Tasks} & \textbf{Third judge calls (of 1{,}500)} & \textbf{Rate (\%)} \\
\midrule
AmbigQA  & 190 & 12.7 \\
FreshQA  & 543 & 36.2 \\
HotpotQA & 120 & 8.0 \\
NQ-Open  & 212 & 14.1 \\
TriviaQA & 253 & 16.9 \\
\midrule
\textbf{Total} & \textbf{1{,}318 / 7{,}500} & \textbf{17.6} \\
\bottomrule
\end{tabular}
\caption{Aggregated third-judge usage under CLEV across tasks. Each task has 
1{,}500 evaluation instances per candidate model (300 examples $\times$ 5 candidate LLMs), 
and the reported counts sum over all five models~\tablefootnote{We also included DeepSeek as our fifth model in the Appendix. See Table~\ref{tab:cost_analysis}.}. Lower values indicate fewer disagreement-triggered escalations.}
\vspace{-0.7em}
\label{tab:clev_agg_escalation}
\end{table}

\subsection{Error analysis}
We randomly sampled 100 error cases (50 false positives and 50 false negatives) from each evaluator to study their behavior. Given EM had 11 false positives, we included all of them in our analysis. 

\noindent\textbf{LLMs are prone to hallucination in justification} where they fabricate reasoning to support their evaluations, and produce detailed but incorrect explanations. In LLM judges, false positives and negatives often result from overlooking critical distinctions between candidate LLM outputs and failing to account for the specificity required by the reference answer. This pattern is particularly noticeable in Mistral 7B, where the model disregards the ground truth and provides evaluations influenced by unknown factors. For example, when evaluating candidate GPT-3.5's response \textit{``The foreign minister of Germany who signed the Treaty of Versailles was Hermann Müller.''} which is correct according to the reference answer \textit{``Hermann Müller''} and human evaluation, Mistral 7B as-a-judge incorrectly marked this response as false and fabricated reasoning in support of its decision.

\noindent\textbf{Specificity bias} occur in some judges. This approach shifts the evaluation towards false negatives by missing semantically similar but structurally different answers. We found many cases when such evaluators failed to account for valid variations in phrasing or granularity, focusing instead on rigid adherence to the reference answer. Compounding these issues are reasoning errors within the evaluators’ own explanations, which often contain overconfident assertions. 

\noindent\textbf{We found several temporal limitations in LLM-based evaluators.} Despite using mostly older datasets with up-to-date models, we observed failures on recent events or evolving contexts. The FreshQA dataset~\citep{vu2023freshllms}, being recent, highlights these temporal limitations. As shown in Table~\ref{tab:freshqa_macro_f1_scores}, LLM judges deviate more from human judgment on FreshQA compared to older datasets like HotpotQA. In dynamic or time-sensitive contexts, they often hallucinate by marking incorrect responses as True. For instance, when asked: \textit{``On what date did the Patriots last play the Miami Dolphins?''}, the evaluator accepted an outdated answer, \textit{``January 1, 2023,''} as correct, ignoring the reference \textit{``November 24, 2024,''} and justifying its mistake. Because multiple judges rely on this same outdated parametric knowledge, FreshQA exhibits substantially higher disagreement between judges, who often converge on answers that were once correct but no longer reflect the current ground-truth.

\begin{table}[t]
\centering
\scriptsize
\begin{tabular}{lccccc}
\toprule
 & \multicolumn{4}{c}{\textbf{Evaluators}} & \\
\cmidrule(lr){2-5}
\textbf{Candid.} & \textbf{Llama} & \textbf{GPT} & \textbf{Mistral} & \textbf{CLEV} & \textbf{Disagr. (\%)} \\
\midrule
Llama     & 0.835 & 0.737 & 0.730 & \cellcolor{green!15}0.917 & 31.3 \\
GPT       & 0.695 & 0.824 & 0.746 & \cellcolor{green!15}0.891 & 44.3 \\
Mixtral   & 0.708 & 0.779 & 0.703 & \cellcolor{green!15}0.936 & 37.3 \\
Mistral   & 0.665 & 0.802 & 0.723 & \cellcolor{green!8}0.880 & 39.7 \\
\bottomrule
\end{tabular}
\caption{Macro-F1 of judges on FreshQA and corresponding disagreement rates. Higher disagreement indicates a more frequent need for a resolving judge (CLEV).}
\vspace{-0.7em}
\label{tab:freshqa_macro_f1_scores}
\end{table}

\section{Related work}
Recent advances in LLMs have unlocked new opportunities for automatic and context-aware evaluation~\citep{li2024llmsasjudgescomp, chianglee2023large, 10.5555/3666122.3668142}. Early work primarily focused on subjective evaluation tasks, such as pairwise comparison for ranking model outputs or single-response scoring for open-ended generation~\citep{chan2024chateval, 10.5555/3666122.3668142, chen2024humansllmsjudgestudy}. These approaches leverage LLMs' strong language understanding and reasoning capabilities to assess qualities like coherence, fluency, and helpfulness. However, such subjective criteria and continuous scoring schemes are less suitable for evaluating objective QA~\citep{badshah2025taletoolaugmented}, where free-form answers are expected to be factually correct with respect to a reference answer~\citep{krumdick2025no, shi2024judgingjudg}.

To address this gap, recent work has explored reference-grounded LLM-as-judge approaches that instruct LLMs to directly verify candidate answers against gold references and output binary True/False verdicts~\citep{kamalloo-etal-2023-evaluating, NEURIPS2023_f323d594}. However, this line of work also highlights key challenges: individual LLM judges can be inconsistent across questions, may over-rely on parametric knowledge when references are incomplete, and often require additional judges or majority voting to stabilize decisions~\citep{khan2024debating}. Furthermore, single judges often lack robustness to adversarial inputs and may hallucinate justifications that appear plausible but are factually incorrect~\citep{hu2024llmbasedeval}.

To mitigate the reliability issues of single-judge systems, multi-judge setups have been explored~\citep{badshah-sajjad-2025-reference, verga2024replacing}. These approaches employ multiple diverse LLM judges and aggregate their decisions, typically through majority voting, to improve reliability and alignment with human evaluation. \citet{badshah-sajjad-2025-reference} demonstrated that using multiple models reduces individual model biases and improves agreement with human judgments. However, these approaches typically rely on fixed ensembles, where all judges are invoked for every instance, regardless of whether the initial judges already agree. This design leads to substantial and unnecessary computational overhead, especially when disagreements are rare. Our work builds on these foundations that invoke a third judge only when needed, achieving the reliability benefits of multi-judge voting while saving compute.

\section{Conclusion}
We studied the challenge of evaluating free-form QA, where traditional metrics fail to capture semantic correctness and single LLM judges may be inconsistent. To address this, we introduced CLEV, a lightweight multi-judge framework that preserves the reliability of majority voting while avoiding its computational cost. By invoking a third judge only when necessary, CLEV achieves substantial to near-perfect agreement with human evaluation while reducing redundant evaluation calls by 80–95\% across tasks. This makes LLM-based evaluation more scalable for large-scale benchmarking. While our study focuses on binary correctness, future work can extend CLEV to capture dimensions such as partial correctness. Looking forward, an important direction is reference-free evaluation, where judges are supported by external verifiers (e.g., retrieval) to assess correctness in settings where explicit references are lacking.

\section{Limitations}
We acknowledge certain limitations in our study: 1) The accuracy of evaluations depends on the quality and clarity of reference answers, which serve as the basis for determining correctness. Incorrect or ambiguous references could affect evaluation outcomes. 2) While we conducted an error analysis of LLM judges and automatic metrics, there may be error cases that were not identified during our manual review, leaving gaps in understanding the full spectrum of evaluation inaccuracies. 3) We acknowledge that our current study is limited to English QA datasets. Extending CLEV to multilingual settings is a promising and important future direction. 4) Since the evaluation is binary, which is standard for free-form QA, signals around partial correctness, fluency, reasoning, and justification are not captured. To tackle this, we also collected detailed rationales that justify each verdict. In our analysis, we leveraged these rationales to capture various insights. We acknowledge the limitation of binary verdicts in representing partial correctness, but we emphasize that this design choice reduces the subjectivity inherent in continuous scoring and promotes clearer agreement between judges. 5) CLEV assumes that primary judges maintain a minimum reliability threshold (based on $\kappa$ and Macro-F1) established during calibration. These thresholds require recalibration when adopting new judge models, domains, or languages. Thus, applying CLEV to new settings may involve additional upfront calibration cost.

\section*{Acknowledgment}
We acknowledge the support of the Natural Sciences and Engineering Research Council of Canada (NSERC), Canada Foundation for Innovation (CFI), and Research Nova Scotia. Advanced computing resources are provided by ACENET, the regional partner in Atlantic Canada, and the Digital Research Alliance of Canada.

\bibliography{anthology,custom}
\bibliographystyle{acl_natbib}

\appendix
\section{Free-form Question-Answering}\label{app:datasets}
In our experiments, we include AmbigQA~\citep{minetal2020ambigqa}, FreshQA~\citep{vu2023freshllms}, HotpotQA~\citep{yang2018hotpotqa}, Natural Questions~\citep{kwiatkowskietal2019}, and TriviaQA~\citep{joshi2017triviaqa}. 
\begin{itemize}
    \item \textbf{AmbigQA}: Focuses on 14K ambiguous questions derived from NQ, requiring systems to identify multiple valid interpretations and generate disambiguated questions alongside corresponding answers. 
    \item \textbf{FreshQA}: A QA benchmark containing 600 questions that consist of a diverse range of types, including those requiring fast-changing world knowledge and questions with false premises that need debunking. It is regularly updated to reflect current information and is designed to evaluate the factual accuracy of LLMs in handling up-to-date and evolving knowledge.
    \item \textbf{HotpotQA}: Contains 113K questions based on Wikipedia. It is designed to test multi-hop reasoning, requiring connections across multiple paragraphs, and includes annotated supporting facts for evaluation.
    \item \textbf{Natural Questions (NQ)}: Consists of real user queries from Google Search, paired with Wikipedia articles. The dataset includes 307K training examples annotated with both long (paragraph) and short (entity-level) answers. 
    \item \textbf{TriviaQA}: Features approximately 650K trivia questions, with evidence sourced from Wikipedia and web searches. These questions often require reasoning across multiple documents for complex answer synthesis.
\end{itemize}

We utilize the validation splits across multiple datasets: the standard validation split for AmbigQA and Natural Questions, the ``distractor'' subset's validation split for HotpotQA, and the ``unfiltered.nocontext'' subset's validation split for TriviaQA. We randomly sampled 300 examples from each dataset using Seed 42.

\section{Baselines}\label{app:baselines}

\paragraph{Exact Match (EM):}For our selected datasets and also free-form QA tasks, EM serves as a standard lexical matching metric to evaluate candidate LLM performance~\citep{izacardgrave2021,105555, gou2024criticllms}. Due to the verbose nature of LLM-generated responses, we adapt EM to classify an answer as correct if any golden answer \(r_i \in R\) appears within the generated response \(\bar{y}\) (i.e., \(r_i \subseteq \bar{y}\)), rather than requiring complete strict string equality (i.e., \(\bar{y} = r_i\)). 

\paragraph{BERTScore:}We use BERTScore~\citep{ZhangKWWA20} which measures similarity by comparing contextualized word embeddings derived from a pre-trained BERT model. This enables the evaluation to focus on semantic correctness rather than exact lexical matches. 
As BERTScore is based on continuous values between -1 and 1, we set a threshold of \(\tau=0.5\) to convert continuous similarity scores into binary 0 and 1. The purpose of this conversion is to allow direct comparison with other evaluation methods. For our implementation, we use the microsoft/deberta-xlarge-mnli\footnote{\url{https://huggingface.co/microsoft/deberta-xlarge-mnli}} model~\citep{he2021deberta}. 

\paragraph{Majority voting}  
This uses three fixed LLM judges to independently evaluate each instance. The final decision is determined by a simple majority across the three verdicts. Unlike CLEV, which selectively invokes the third judge only in cases of disagreement, this method uniformly engages all judges, leading to a higher computational cost.


\section{Human evaluation}\label{app:human_evaluation}
This section provides detailed guidelines for human annotators responsible for evaluating the outputs of candidate LLMs. The goal is to ensure consistency and objectivity across all evaluations~\citep{yu-etal-2024-latent}. These guidelines provide clear instructions for assessing each model's response based on its alignment with the reference answer and contextual relevance.

\begin{table*}[htbp]
\centering
\begin{tabular}{lccccc}
\toprule
\textbf{LLMs} & \textbf{AmbigQA} & \textbf{FreshQA} & \textbf{HotpotQA} & \textbf{NQ-Open} & \textbf{TriviaQA} \\
\midrule
DeepSeek & 0.975 & 0.949 & 0.986 & 0.889 & 0.456 (\(\kappa\) paradox) \\
Llama    & 0.945 & 0.962 & 0.973 & 0.985 & 0.935 \\
GPT      & 0.989 & 0.973 & 0.982 & 0.990 & 0.948 \\
Mixtral  & 0.981 & 0.945 & 0.996 & 0.977 & 0.936 \\
Mistral  & 0.978 & 0.932 & 0.981 & 0.978 & 0.975 \\
\bottomrule
\end{tabular}
\caption{Fleiss' Kappa scores of human annotators across models and tasks.}
\label{tab:fleiss-kappa_r}
\end{table*}

\begin{table*}[htbp]
\centering
\begin{tabular}{lccccc}
\toprule
\textbf{LLMs} & \textbf{AmbigQA} & \textbf{FreshQA} & \textbf{HotpotQA} & \textbf{NQ-Open} & \textbf{TriviaQA} \\
\midrule
DeepSeek  & 99.0\% & 98.0\% & 99.7\% & 92.0\% & 90.0\%    \\
Llama     & 96.3\% & 98.0\%    & 98.0\% & 99.0\% & 99.0\% \\
GPT       & 99.3\% & 99.3\%    & 98.7\% & 99.3\% & 99.0\% \\
Mixtral   & 98.7\% & 98.0\%    & 99.7\% & 98.3\% & 98.3\% \\
Mistral   & 98.3\% & 97.0\%    & 98.7\% & 98.3\% & 99.0\% \\
\bottomrule
\end{tabular}
\caption{Human annotators percent agreement scores across candidate models and tasks.}
\label{tab:percent_agreement_human}
\end{table*}

\subsection{Guidelines}
Dear Evaluator, 

\noindent Thank you for your valuable contribution to this evaluation process. These guidelines outline the process for evaluating Large Language Model (LLM) outputs for the given tasks.  As annotators, you will receive three components for each evaluation instance: the input question, reference answer(s), and the model's response. Your task is to evaluate the responses independently and score them on a binary scale: `1' for `True' (correct) and `0' for `False' (incorrect). \\

\noindent A response warrants a score of `1' when it demonstrates semantic equivalence with the reference answer, even if expressed through alternative phrasing or structure. This includes acceptable variations such as synonym usage and structural variations. Additional contextual information is acceptable as long as it doesn't introduce errors. \\

\noindent Responses receive a score of `0' when they contain factual errors, miss crucial elements from the reference answer, or demonstrate contextual misalignment. Partial answers that omit essential information should be marked incorrect, regardless of the accuracy of included content. When multiple reference answers are provided, a response is correct if it fully aligns with at least one reference. \\

\noindent You are encouraged to use internet resources when needed to verify specific facts, terminology, or potential synonyms that may affect your evaluation decision. However, the reference answer should remain the primary basis for evaluation. Focus on whether the model's response conveys the same core information as the reference answer. To maintain reliability, document any challenging cases requiring further discussion with other annotators. \\

\section{Evaluation Metrics}
We compute \textbf{Fleiss' Kappa} (\(\kappa\))~\citep{fleiss1973equivalence} and \textbf{percent agreement} to assess inter-rater reliability among human annotators. Similarly, we use \textbf{Cohen's kappa}~\citep{mchugh2012interrater} to find the agreement between each evaluator and the human majority to obtain instance-level comparison. Due to the high-class imbalance in TriviaQA, kappa scores can be misleadingly low despite high raw agreement - a known limitation called the \textit{``kappa paradox''}~\citep{cicchetti1990high}. Therefore, we treat the evaluation as a binary classification task where we consider each evaluator's predictions against the human majority and report \textbf{Macro-F1} scores which give equal weight to both classes regardless of their frequency in the selected random samples. 

To quantify the efficiency of our approach, we report the \textbf{disagreement rate} between the two primary judges that indicates how often the third model is required, thereby revealing the reduction in third-model usage compared to always employing three judges. Formally,\vspace{-8pt}

{\small
\[
\text{Disagreement rate (\%)} = \left( \frac{1}{N} \sum_{i=1}^{N} \mathbb{I}[V_{i_1} \neq V_{i_2}] \right) \times 100
\]
}

\noindent where $N$ is the total number of evaluation instances and $\mathbb{I}[\cdot]$ is the indicator function that equals 1 when the condition is satisfied and 0 otherwise.

\subsection{Inter-human annotator agreement}
We calculate Fleiss' Kappa (\(\kappa\))~\citep{fleiss1973equivalence} and percent agreement to assess inter-rater reliability among human annotators.

Fleiss’ Kappa is defined as:
\[
\kappa = \frac{\bar{P} - P_e}{1 - P_e},
\]
where $\bar{P}$ is the average observed agreement among annotators, and $P_e$ is the expected agreement by chance.

Percent agreement is calculated as:
\[
\text{Percent Agreement} = \left(\frac{\text{Agreements}}{\text{Total Annotations}}\right) \times 100
\]

Table~\ref{tab:fleiss-kappa_r} and~\ref{tab:percent_agreement_human} show the inter-annotator agreement across models and tasks. The results demonstrate high reliability, with Fleiss' Kappa scores consistently above 0.93 for most tasks. The highest agreement is observed in Mixtral evaluations on HotpotQA (\(\kappa = 0.996\)), and GPT on NQ-Open (\(\kappa = 0.990\)). In FreshQA, which shows lower Kappa scores, the agreement among annotators remains high including 99.3\% in GPT and 98.0\% in Mixtral.

The percent agreement scores in Table~\ref{tab:percent_agreement_human} further confirm strong inter-annotator consistency. Most models achieve over 98\% agreement across AmbigQA, HotpotQA, NQ-Open, and TriviaQA. However, DeepSeek exhibits lower agreement on NQ-Open (92.0\%) and TriviaQA (90.0\%). This indicates a variance in human ratings for these tasks.

\section{Additional results}\label{app:additional_results}
This section provides further results and analysis of conventional metrics and LLM-based evaluators. 

As evidenced by high Cohen's kappa and Macro F1 scores in Table~\ref{app:kappa_scores_evaluators_n} and~\ref{app:macro_f1_scores}, CLEV maintains a strong alignment with human evaluation. This represents a substantial improvement over individual model performance, where individual judges generally showed varying levels of agreement with human evaluation. Overall, LLM-as-a-judge works better with larger models. This is particularly noticeable in Llama and GPT, which achieve greater performance across AmbigQA, HotpotQA, and NQ-Open compared to smaller models. This indicates an important scaling law in evaluation capability~\citep{kaplan2020scalinglawsneurallanguage, 10.5555/3666122.3668142, openai2024gpt4technicalreport}. However, we also found that the most advanced models are not always guaranteed to be the best evaluators. We observed slightly comparable performance through the small open-source Mistral7B. For instance, when evaluating candidate Mixtral 8x7B on AmbigQA (see Table~~\ref{app:macro_f1_scores}), Mistral 7B as-a-judge outperformed (0.944) judge GPT-3.5-turbo (0.891). Regardless, we observe relatively lower Macro-F1 scores for all LLM judges in TriviaQA. 

\begin{table}[t]
\centering
\scriptsize
\setlength{\tabcolsep}{4pt}
\resizebox{\linewidth}{!}{
\begin{tabular}{llccccccc} 
\toprule
 & & \multicolumn{7}{c}{\textbf{Evaluators}} \\ 
\cmidrule(lr){3-9} 
\textbf{LLMs} & \textbf{Tasks} & \textbf{EM} & \textbf{BS} & 
\textbf{Llama} & \textbf{GPT} & \textbf{Mixtral} & \textbf{Mistral} & \textbf{CLEV} \\
\midrule
\multirow{5}{*}{Llama} 
& AmbigQA  & 0.518 & 0.283 & 
0.888 & 0.844 & 0.824 & 0.858 & \cellcolor{green!15}0.911 \\
& HotpotQA & 0.577 & 0.498 & 
0.877 & 0.899 & 0.820 & 0.832 & \cellcolor{green!15}0.953 \\
& NQ-Open  & 0.381 & 0.437 & 
0.833 & 0.793 & 0.816 & 0.738 & \cellcolor{green!15}0.927 \\
& TriviaQA & 0.281 & 0.564 & 
0.547 & 0.439 & 0.396 & 0.299 & \cellcolor{green!8}0.684 \\
\midrule
\multirow{5}{*}{GPT} 
& AmbigQA  & 0.561 & 0.252 & 
0.944 & 0.897 & 0.861 & 0.853 & \cellcolor{green!15}0.967 \\
& HotpotQA & 0.604 & 0.300 & 
0.953 & 0.973 & 0.873 & 0.933 & \cellcolor{green!15}0.987 \\
& NQ-Open  & 0.453 & 0.218 & 
0.884 & 0.824 & 0.824 & 0.829 & \cellcolor{green!15}0.956 \\
& TriviaQA & 0.335 & 0.364 & 
0.650 & 0.401 & 0.580 & 0.467 & \cellcolor{green!8}0.775 \\
\midrule
\multirow{5}{*}{Mixtral} 
& AmbigQA  & 0.546 & 0.337 & 
0.896 & 0.781 & 0.909 & 0.887 & \cellcolor{green!15}0.951 \\
& HotpotQA & 0.546 & 0.349 & 
0.940 & 0.933 & 0.859 & 0.940 & \cellcolor{green!15}0.973 \\
& NQ-Open  & 0.371 & 0.301 & 
0.879 & 0.728 & 0.899 & 0.815 & \cellcolor{green!15}0.913 \\
& TriviaQA & 0.317 & 0.390 & 
0.625 & 0.605 & 0.678 & 0.436 & \cellcolor{green!8}0.764 \\
\midrule
\multirow{5}{*}{Mistral} 
& AmbigQA  & 0.599 & 0.254 & 
0.893 & 0.893 & 0.893 & 0.860 & \cellcolor{green!15}0.953 \\
& HotpotQA & 0.605 & 0.383 & 
0.937 & 0.902 & 0.895 & 0.937 & \cellcolor{green!15}0.958 \\
& NQ-Open  & 0.484 & 0.291 & 
0.851 & 0.838 & 0.878 & 0.840 & \cellcolor{green!15}0.953 \\
& TriviaQA & 0.467 & 0.239 & 
0.758 & 0.725 & 0.645 & 0.470 & \cellcolor{green!8}0.854 \\
\bottomrule
\end{tabular}
}
\caption{Cohen's Kappa scores displaying the agreement levels of individual and multiple (CLEV) evaluators with human judgments across candidate models and tasks.
}
\label{app:kappa_scores_evaluators_n}
\end{table}

\begin{table}[t]
\centering
\scriptsize
\setlength{\tabcolsep}{4pt}
\resizebox{\linewidth}{!}{
\begin{tabular}{llccccccc} 
\toprule
 & & \multicolumn{7}{c}{\textbf{Evaluators}} \\ 
\cmidrule(lr){3-9} 
\textbf{LLMs} & \textbf{Tasks} & \textbf{EM} & \textbf{BS} & 
\textbf{Llama} & \textbf{GPT} & \textbf{Mixtral} & \textbf{Mistral} & \textbf{CLEV} \\
\midrule
\multirow{5}{*}{Llama} 
& AmbigQA  & 0.744 & 0.641 & 
0.944 & 0.922 & 0.912 & 0.929 & \cellcolor{green!15}0.955 \\
& HotpotQA & 0.778 & 0.745 & 
0.939 & 0.949 & 0.910 & 0.916 & \cellcolor{green!15}0.976 \\
& NQ-Open  & 0.653 & 0.718 & 
0.916 & 0.896 & 0.907 & 0.869 & \cellcolor{green!15}0.964 \\
& TriviaQA & 0.612 & 0.782 & 
0.772 & 0.717 & 0.695 & 0.640 & \cellcolor{green!8}0.842 \\
\midrule
\multirow{5}{*}{GPT}
& AmbigQA  & 0.792 & 0.622 & 
0.972 & 0.949 & 0.930 & 0.927 & \cellcolor{green!15}0.984 \\
& HotpotQA & 0.794 & 0.623 & 
0.977 & 0.987 & 0.936 & 0.966 & \cellcolor{green!15}0.993 \\
& NQ-Open  & 0.703 & 0.606 & 
0.942 & 0.911 & 0.911 & 0.914 & \cellcolor{green!15}0.978 \\
& TriviaQA & 0.646 & 0.681 & 
0.824 & 0.700 & 0.789 & 0.730 & \cellcolor{green!9}0.887 \\
\midrule
\multirow{5}{*}{Mixtral}
& AmbigQA  & 0.760 & 0.666 & 
0.948 & 0.891 & 0.955 & 0.944 & \cellcolor{green!15}0.975 \\
& HotpotQA & 0.761 & 0.657 & 
0.970 & 0.966 & 0.930 & 0.970 & \cellcolor{green!15}0.987 \\
& NQ-Open  & 0.650 & 0.649 & 
0.939 & 0.863 & 0.950 & 0.908 & \cellcolor{green!15}0.956 \\
& TriviaQA & 0.625 & 0.695 & 
0.812 & 0.803 & 0.838 & 0.716 & \cellcolor{green!8}0.882 \\
\midrule
\multirow{5}{*}{Mistral}
& AmbigQA  & 0.792 & 0.622 & 
0.947 & 0.947 & 0.947 & 0.930 & \cellcolor{green!15}0.977 \\
& HotpotQA & 0.796 & 0.673 & 
0.969 & 0.951 & 0.947 & 0.969 & \cellcolor{green!15}0.979 \\
& NQ-Open  & 0.726 & 0.639 & 
0.925 & 0.919 & 0.939 & 0.920 & \cellcolor{green!15}0.976 \\
& TriviaQA & 0.718 & 0.608 & 
0.879 & 0.863 & 0.822 & 0.735 & \cellcolor{green!15}0.927 \\
\bottomrule
\end{tabular}
}
\caption{Macro-F1 scores of individual and multiple (CLEV) evaluators applied to different candidate LLMs and associated tasks.
}
\label{app:macro_f1_scores}
\end{table}

Unlike lexical matching and neural-based metrics, each LLM-as-a-judge shows overall performance close to the human majority. The proposed CLEV method consistently achieves comparable or slightly better alignment with the human majority. Conventional metrics such as EM severely underestimate the candidate LLMs' performance. Contrarily, BERTScore tends to overestimate the performance except in some cases such as when evaluating Llama on AmbigQA and NQ-Open. 

EM underestimates performance because it requires a candidate's response to exactly match one of the reference answers. This rigid, lexical approach fails to account for valid paraphrases, synonyms, or alternative expressions that convey the same meaning. In free-form QA tasks, where there can be multiple correct answers phrased in various ways, EM's strict criteria often penalize responses that are semantically accurate but differ slightly in wording. As a result, it underestimates the true capabilities of candidate LLMs, leading to an incomplete assessment of their performance.

BERTScore relies on token-level semantic similarity, which rewards shallow lexical overlap rather than actual factual accuracy. For example, in cases where minor differences in wording (e.g., ``The Treaty of Versailles was signed in 1919.'' versus ``The Treaty of Versailles ended in 1919.'') lead to opposing factual claims, BERTScore still scores the response high due to its emphasis on matching tokens (e.g., ``signed'' versus ``ended''). Additionally, verbosity bias and threshold instability further inflate its raw accuracy (see Table~\ref{tab:task_oriented_llm_scores_free}). However, when comparing raw accuracy with instance-level agreement metrics like Cohen’s kappa, which adjusts for class imbalance and penalizes asymmetric errors, the limitations of BERTScore become apparent.

\subsection{Cost analysis}
To assess the efficiency of CLEV, we track the number of times the third judge is invoked, which directly corresponds to disagreement between the two primary judges. As shown in Table~\ref{tab:cost_analysis}, across 7,500 evaluation instances, the third judge is required only 1,318 times, representing just 17.6\% of the total cases. This implies an 82.4\% reduction in third-judge usage compared to a full majority-voting setup, where every instance would involve all three models.

Disagreement rates vary across tasks and models. For example, GPT shows only 5.7\% disagreement on HotpotQA, while FreshQA exhibits higher disagreement (up to 44.3\%) for some judge combinations. This behavior allows CLEV to scale efficiently: it concentrates computational effort only where model uncertainty exists, minimizing redundant inference. In contrast to fixed-cost evaluation schemes, CLEV offers a cost-efficient alternative that maintains high evaluation quality while significantly reducing compute usage.

\begin{table*}[ht]
\centering
\begin{tabular}{lcccc}
\toprule
\textbf{Candidate LLMs} & \textbf{Tasks} & \textbf{Samples} & \textbf{Disagreement Rates (\%)} & \textbf{Third Judge Usage} \\
\midrule
DeepSeek & AmbigQA  & 300 & 25.7 & 77  \\
  & FreshQA  & 300 & 28.3 & 85  \\
  & HotpotQA & 300 & 10.7 & 32  \\
  & NQ-Open  & 300 & 12.0 & 36  \\
  & TriviaQA & 300 & 14.3 & 43  \\
\midrule
Llama  & AmbigQA  & 300 & 10.0 & 30  \\
  & FreshQA  & 300 & 31.3 & 94  \\
  & HotpotQA & 300 & 13.0 & 39  \\
  & NQ-Open  & 300 & 18.0 & 54  \\
  & TriviaQA & 300 & 17.0 & 51  \\
\midrule
GPT  & AmbigQA  & 300 & 7.0  & 21  \\
 & FreshQA  & 300 & 44.3 & 133 \\
  & HotpotQA & 300 & 5.7  & 17  \\
  & NQ-Open  & 300 & 13.0 & 39  \\
  & TriviaQA & 300 & 15.7 & 47  \\
\midrule
Mixtral  & AmbigQA  & 300 & 9.0  & 27  \\
  & FreshQA  & 300 & 37.3 & 112 \\
  & HotpotQA & 300 & 4.7  & 14  \\
  & NQ-Open  & 300 & 13.0 & 39  \\
  & TriviaQA & 300 & 17.0 & 51  \\
\midrule
Mistral  & AmbigQA  & 300 & 11.7 & 35  \\
  & FreshQA  & 300 & 39.7 & 119 \\
  & HotpotQA & 300 & 6.0  & 18  \\
  & NQ-Open  & 300 & 14.7 & 44  \\
  & TriviaQA & 300 & 20.3 & 61  \\
\midrule
\textbf{Total}  &   & \textbf{7500}  &  & \textbf{1318}  \\
\bottomrule
\end{tabular}
\caption{Cost-efficiency analysis of CLEV: Summary of disagreement rates and third judge usage across candidate models and tasks}
\label{tab:cost_analysis}
\end{table*}

\subsection{DeepSeek as the third judge}
To assess the impact of using DeepSeek as the third judge in CLEV, we conducted experiments by replacing GPT-3.5-turbo with  DeepSeek-R1~\citep{deepseekai2025deepseekr1}. We evaluated this setup using candidate GPT-3.5 on TriviaQA, candidate DeepSeek on NQ-Open, and candidate Llama on FreshQA. The primary judges remained Llama and Mistral, and the third judge (i.e., DeepSeek-R1) is invoked only in cases of disagreement. Our findings indicate that DeepSeek, as the third judge, achieves strong performance, with Macro-F1 scores of 91.23 on TriviaQA, 79.11 on NQ-Open, and 0.914 on FreshQA.

\subsection{Evaluating with GPT-4o as-a-judge}
While a single state-of-the-art evaluator can achieve strong performance in many cases, the dual-LLM framework remains critical for ensuring robustness, particularly in high-stakes or ambiguous scenarios.

To explore the potential of a more powerful single LLM, we evaluated GPT-3.5-turbo on HotpotQA and TriviaQA using GPT-4o as a judge. With this configuration, GPT-4o as the evaluator achieved a Macro-F1 score of 0.946 on HotpotQA, demonstrating its exceptional capability. However, the same GPT-4o judge achieved only 0.784 on TriviaQA, which falls short of CLEV’s performance of 0.887. This shows that even the most advanced models show inconsistencies when evaluating free-form QA. This is particularly critical in precision-sensitive domains where minor errors can have outsized consequences.

In such settings, CLEV’s ensemble approach acts as a safeguard. When employing CLEV with GPT-3.5-turbo as the third judge, we achieved an even higher Macro-F1 of 0.984 on HotpotQA, surpassing the performance of a single GPT-4o. Interestingly, when we experimented with DeepSeek as the third judge in CLEV, performance remained strong at 0.963 Macro-F1, indicating that CLEV's benefits are not solely tied to a specific third judge model.

\subsection{Majority voting-based evaluation}
We conducted additional experiments utilizing a traditional majority voting approach for evaluating candidate LLMs performance. Given \( n \) annotators and a binary classification, the majority label is defined as:

\[
y_{\text{majority}} = \begin{cases} 
1 & \text{if } \sum_{i=1}^{n} y_i > \frac{n}{2}, \\
0 & \text{otherwise},
\end{cases}
\]

where \( y_i \) represents the label assigned by the \( i \)th annotator.

In this setup, we employed three LLM judges of equal weight: Llama, GPT-3.5, and Mistral to evaluate candidate models' generated responses. For every evaluation instance, each judge provided an independent binary verdict (True or False). The final decision is determined through a simple majority vote across these three verdicts.

As presented in Table~\ref{tab:majvote}, CLEV matches or closely approaches the Macro F1 and Cohen’s Kappa scores of the three-judge majority across almost all tasks and candidate LLMs. For example, on HotpotQA, evaluating candidate Llama with CLEV achieves a Macro F1 of 97.6\% (compared to 97.6\% for majority voting) and a Cohen’s Kappa of 0.95, while for GPT-3.5 on AmbigQA, CLEV reaches a Macro F1 of 98.4\% (versus 98.3\% for majority voting), indicating a negligible performance difference. Even in high-disagreement tasks like TriviaQA, where the primary judges (e.g., Mistral) disagree 20.3\% of the time, CLEV retains strong alignment (with a Macro F1 of 92.7 compared to 93.5 for majority voting). Minor deviations, such as the one observed for candidate Mixtral on TriviaQA (CLEV’s Macro F1 = 0.88 vs. 0.95 for majority voting), reflect rare instances where both the primary judges and the third judge make errors, yet these outliers are substantially outweighed by the computational savings offered by the selective third judge.

\begin{table*}[t]
\centering
\begin{tabular}{llccccc}
\toprule
\multirow{2}{*}{\textbf{Candidate LLM}} & \multirow{2}{*}{\textbf{Task}} & \multicolumn{2}{c}{\textbf{Majority Voting}} & \multirow{2}{*}{\textbf{Disagreement (\%)}} & \multicolumn{2}{c}{\textbf{CLEV}} \\
\cmidrule(lr){3-4}\cmidrule(lr){6-7}
 &  & \textbf{Macro F1} & \textbf{Kappa} &  & \textbf{Macro F1} & \textbf{Kappa} \\
\midrule
\multirow{4}{*}{Llama} 
  & AmbigQA  & 95.5 & 0.91 & 10.0 & 95.5 & 0.91 \\
  & HotpotQA & 97.6 & 0.95 & 13.0 & 97.6 & 0.95 \\
  & NQ-Open  & 96.3 & 0.93 & 18.0 & 96.4 & 0.92 \\
  & TriviaQA & 84.1 & 0.68 & 17.0 & 84.2 & 0.68 \\
\midrule
\multirow{4}{*}{GPT} 
  & AmbigQA  & 98.3 & 0.97 & 7.0  & 98.4 & 0.96 \\
  & HotpotQA & 99.3 & 0.99 & 5.7  & 99.3 & 0.98 \\
  & NQ-Open  & 97.8 & 0.96 & 13.0 & 97.8 & 0.95 \\
  & TriviaQA & 90.5 & 0.81 & 15.7 & 88.7 & 0.77 \\
\midrule
\multirow{4}{*}{Mixtral} 
  & AmbigQA  & 98.9 & 0.98 & 9.0  & 97.5 & 0.95 \\
  & HotpotQA & 98.6 & 0.97 & 4.7  & 98.7 & 0.97 \\
  & NQ-Open  & 98.3 & 0.97 & 13.0 & 95.6 & 0.91 \\
  & TriviaQA & 95.0 & 0.90 & 17.0 & 88.2 & 0.76 \\
\midrule
\multirow{4}{*}{Mistral} 
  & AmbigQA  & 97.6 & 0.95 & 11.7 & 97.7 & 0.95 \\
  & HotpotQA & 97.9 & 0.96 & 6.0  & 97.9 & 0.95 \\
  & NQ-Open  & 97.6 & 0.95 & 14.7 & 97.6 & 0.95 \\
  & TriviaQA & 93.5 & 0.87 & 20.3 & 92.7 & 0.85 \\
\bottomrule
\end{tabular}
\caption{Comparison between Majority Voting (Llama+GPT-3.5+Mistral) and CLEV (GPT-3.5 as the third judge). For each candidate LLM and task, the table reports Macro F1 and Cohen’s Kappa scores under Majority Voting, the disagreement rate (in \%), and the corresponding scores using CLEV. Small differences in performance between the majority voting setup and CLEV are due to randomness (see Section~\ref{app:consistency_jug}).}
\label{tab:majvote}
\end{table*}

\subsection{Impact of prompt variations}
The effectiveness and consistency of LLM-based evaluation are significantly influenced by prompt design. Variations in prompt structure, reasoning order, explanation requirements, and task-specific examples can lead to notable differences in model verdicts. To analyze the robustness of the LLM judges in free-form QA, we conducted ablation studies on different prompt variations using Mistral as the candidate model and GPT as the judge.

\subsubsection{Consistency in judgment across multiple trials}\label{app:consistency_jug}
LLMs generate random text even at a temperature of 0. To assess whether this affects evaluation consistency, we repeated the same evaluation task five times for 300 Mistral-generated responses for HotpotQA. In this ablation study, we use GPT-3.5 as a judge. 

{\begin{itemize}
    \item \textbf{Verdict stability:} GPT produced identical True/False verdicts in 98\% of cases, with minor variations observed in only 1–2 instances. This suggest that its binary decision-making process remains largely stable even across multiple trials.
    \item \textbf{Explanation variability:} While verdicts remained consistent, the rationales and explanations provided by GPT across trials, often cited different supporting facts for the same judgment.
\end{itemize}

\subsubsection{Few-shot vs. zero-shot prompting}
We investigated the impact of few-shot prompting where we included three \textbf{task-specific examples} in the prompt to guide the judge's decision-making process. We found that adding few-shot examples resulted in a 2\% increase in Macro-F1 scores. However, few-shot prompting introduced rigid decision patterns—the model sometimes over-applied reasoning from the examples rather than adapting flexibly to novel cases. For instance, multi-hop reasoning cases from HotpotQA, the judge model consistently followed the structure of the provided examples, even when the correct reasoning required a different approach.}

\subsubsection{Explanation requirement: Binary verdict vs. justification-based evaluation}
To test whether requiring the model to generate explanations alongside verdicts improves judgment reliability, we compared two settings:

\begin{itemize}
    \item \textbf{Binary verdict-only evaluation:} The model was instructed to provide only a True/False response without any explanation.
    \item \textbf{Justification-based evaluation:} The model was required to explain its reasoning before delivering the final verdict.
\end{itemize}
\noindent We found that:
\begin{itemize}
    \item \textbf{Higher verdict volatility in verdict-only mode:} When explanations were removed, 13\% of verdicts changed between repeated evaluations of the same responses.
    \item \textbf{Reduced alignment with human judgment:} Cohen’s Kappa agreement with human annotators dropped from 0.95 to 0.72, highlighting that rationale-based prompts lead to more stable and accurate decisions.
\end{itemize}

\subsubsection{Reason-first vs. verdict-first prompting}
In the verdict-first approach, the model is instructed to provide a True/False answer before justifying its decision, whereas in the reason-first approach, the model is asked to generate reasoning first and then conclude with a verdict. Experimental results showed no significant difference in accuracy or agreement scores between these two formats. 

\subsection{G-Eval: reference-free evaluation of free-form question-answering}\label{app:reference_free}
Existing LLM-based evaluators such as G-Eval~\citep{liuetal2023g} are designed for reference-free, subjective tasks (e.g., summarization, dialogue), where evaluation criteria (e.g., coherence, fluency) are inherently ambiguous and scored on Likert scales. These frameworks prioritize qualitative judgments rather than binary factual correctness. In contrast, CLEV is explicitly tailored for reference-dependent, objective evaluation in free-form QA, where answers are either factually correct or incorrect based on alignment with explicit ground-truth references.

To validate this distinction, we tailored the G-eval~\citep{liuetal2023g} to investigate the capability of LLM-as-a-judge in reference-free settings. In this setting, we modify the evaluation prompt by excluding the reference answer \( r \) and directly prompt the evaluator model as \( P = \{x, \bar{y}\} \) along with instructions such as correctness.

The performance of LLM-as-a-judge drastically changes in reference-free settings. Without access to the ground truth references, we observe a stark decline in evaluation capability across all models (see Table~\ref{tab:task_oriented_llm_scores_free} and~\ref{tab:macro_f1_scores_free} \textcolor{blue}{values in blue}). This systematic deterioration spans all tasks and model combinations, though its severity varies by context. HotpotQA and NQ-Open, with their demands for complex reasoning, exemplify this challenge most clearly. The substantial gap between reference-based and reference-free evaluation underscores the crucial role of reference answers in reliable assessment.

\begin{table*}[t]
\setlength{\tabcolsep}{6pt}
\scriptsize
\centering
\begin{tabular}{p{6em}lcccccccc}
\toprule
\multicolumn{1}{l}{\textbf{Candidate LLMs}} & \textbf{Tasks} & \multicolumn{7}{c}{\textbf{Evaluators}} \\
\cmidrule(lr){3-9}
& & \textbf{EM} & \textbf{BERTScore} & \textbf{Human Majority} & \textbf{Llama-3.1-70B} & \textbf{GPT-3.5-turbo} & \textbf{Mixtral-8x7B} & \textbf{Mistral-7B} \\
\midrule
\multirow{4}{7em}{Llama-3.1-70B} & AmbigQA & 42.3 & 63.0 & 67.0 & 65.3 [\textcolor{blue}{83.3}] & 64.7 [\textcolor{blue}{84.7}] & 63.0 [\textcolor{blue}{76.0}] & 66.0 [\textcolor{blue}{80.3}] \\
& HotpotQA & 34.3 & 67.7 & 56.3 & 58.3 [\textcolor{blue}{81.0}] & 54.0 [\textcolor{blue}{81.0}] & 50.7 [\textcolor{blue}{67.3}] & 52.7 [\textcolor{blue}{69.3}] \\
& NQ-Open & 31.7 & 61.7 & 66.3 & 62.7 [\textcolor{blue}{89.0}] & 60.0 [\textcolor{blue}{89.3}] & 59.0 [\textcolor{blue}{81.0}] & 66.7 [\textcolor{blue}{81.0}] \\
& TriviaQA & 74.3 & 94.0 & 94.7 & 90.3 [\textcolor{blue}{90.3}] & 90.0 [\textcolor{blue}{90.3}] & 88.7 [\textcolor{blue}{89.0}] & 84.7 [\textcolor{blue}{84.0}] \\
\midrule
\multirow{4}{6em}{GPT-3.5} & AmbigQA & 49.7 & 78.0 & 71.7 & 70.0 [\textcolor{blue}{79.0}] & 68.0 [\textcolor{blue}{81.0}] & 65.7 [\textcolor{blue}{79.0}] & 71.0 [\textcolor{blue}{84.3}] \\
& HotpotQA & 33.7 & 80.0 & 54.0 & 53.0 [\textcolor{blue}{85.3}] & 52.7 [\textcolor{blue}{85.7}] & 51.7 [\textcolor{blue}{82.3}] & 54.0 [\textcolor{blue}{86.3}] \\
& NQ-Open & 36.3 & 74.0 & 65.3 & 62.7 [\textcolor{blue}{83.7}] & 59.0 [\textcolor{blue}{90.7}] & 59.0 [\textcolor{blue}{87.0}] & 67.0 [\textcolor{blue}{89.7}] \\
& TriviaQA & 74.3 & 95.3 & 93.0 & 89.3 [\textcolor{blue}{89.0}] & 90.7 [\textcolor{blue}{88.7}] & 89.7 [\textcolor{blue}{90.3}] & 86.3 [\textcolor{blue}{84.3}] \\
\midrule
\multirow{4}{6em}{Mixtral-8x7B} & AmbigQA & 37.7 & 70.3 & 61.7 & 57.3 [\textcolor{blue}{74.7}] & 62.0 [\textcolor{blue}{82.3}] & 59.3 [\textcolor{blue}{79.7}] & 61.7 [\textcolor{blue}{80.7}] \\
& HotpotQA & 25.0 & 69.7 & 47.0 & 45.3 [\textcolor{blue}{80.0}] & 45.7 [\textcolor{blue}{84.7}] & 44.7 [\textcolor{blue}{72.0}] & 46.0 [\textcolor{blue}{78.0}] \\
& NQ-Open & 23.7 & 63.7 & 56.7 & 52.7 [\textcolor{blue}{81.7}] & 47.7 [\textcolor{blue}{90.3}] & 52.3 [\textcolor{blue}{85.7}] & 59.7 [\textcolor{blue}{89.7}] \\
& TriviaQA & 64.7 & 91.3 & 90.7 & 86.3 [\textcolor{blue}{85.7}] & 89.7 [\textcolor{blue}{89.0}] & 86.0 [\textcolor{blue}{86.7}] & 85.3 [\textcolor{blue}{86.0}] \\
\midrule
\multirow{4}{6em}{Mistral-7B} & AmbigQA & 31.0 & 61.7 & 49.7 & 46.3 [\textcolor{blue}{61.0}] & 47.7 [\textcolor{blue}{78.7}] & 46.3 [\textcolor{blue}{74.7}] & 53.3 [\textcolor{blue}{85.0}] \\
& HotpotQA & 23.7 & 64.7 & 40.0 & 39.0 [\textcolor{blue}{64.3}] & 38.0 [\textcolor{blue}{83.3}] & 37.0 [\textcolor{blue}{62.0}] & 39.0 [\textcolor{blue}{77.0}] \\
& NQ-Open & 22.7 & 60.0 & 46.0 & 40.0 [\textcolor{blue}{72.3}] & 43.3 [\textcolor{blue}{85.7}] & 41.3 [\textcolor{blue}{78.0}] & 50.0 [\textcolor{blue}{92.3}] \\
& TriviaQA & 62.0 & 94.3 & 83.7 & 81.3 [\textcolor{blue}{80.7}] & 81.0 [\textcolor{blue}{81.0}] & 79.7 [\textcolor{blue}{80.7}] & 85.0 [\textcolor{blue}{84.7}] \\
\bottomrule
\end{tabular}
\caption{Overall performance (Raw Accuracy) of candidate LLMs across free-form QA tasks. Values [in \textcolor{blue}{blue}] represent LLM-as-a-judge in the reference-free mood.}

\label{tab:task_oriented_llm_scores_free}
\end{table*}

\begin{table*}[t]
\setlength{\tabcolsep}{6pt}
\scriptsize
\centering
\begin{tabular}{p{6em}lccccccc}
\toprule
\multicolumn{1}{l}{\textbf{Candidate LLMs}} & \textbf{Tasks} & \multicolumn{7}{c}{\textbf{Evaluators}} \\
\cmidrule(lr){3-9}
& & \textbf{EM} & \textbf{BERTScore} & \textbf{Llama-3.1-70B} & \textbf{GPT-3.5-turbo} & \textbf{Mixtral-8x7B} & \textbf{Mistral-7B} & \textbf{CLEV} \\
\midrule
\multirow{4}{7em}{Llama-3.1-70B} 
& AmbigQA  & 0.744 & 0.641 & 0.944 [\textcolor{blue}{0.629}] & 0.922 [\textcolor{blue}{0.604}] & 0.912 [\textcolor{blue}{0.669}] & 0.929 [\textcolor{blue}{0.631}] & 0.955 [\textcolor{blue}{0.637}] \\
& HotpotQA & 0.778 & 0.745 & 0.939 [\textcolor{blue}{0.628}] & 0.949 [\textcolor{blue}{0.574}] & 0.910 [\textcolor{blue}{0.665}] & 0.916 [\textcolor{blue}{0.640}] & 0.976 [\textcolor{blue}{0.623}] \\
& NQ-Open  & 0.653 & 0.718 & 0.916 [\textcolor{blue}{0.606}] & 0.896 [\textcolor{blue}{0.560}] & 0.907 [\textcolor{blue}{0.639}] & 0.869 [\textcolor{blue}{0.622}] & 0.964 [\textcolor{blue}{0.610}] \\
& TriviaQA & 0.612 & 0.782 & 0.772 [\textcolor{blue}{0.772}] & 0.717 [\textcolor{blue}{0.628}] & 0.695 [\textcolor{blue}{0.678}] & 0.640 [\textcolor{blue}{0.633}] & 0.842 [\textcolor{blue}{0.747}] \\
\midrule
\multirow{4}{6em}{GPT-3.5} 
& AmbigQA  & 0.792 & 0.622 & 0.972 [\textcolor{blue}{0.686}] & 0.949 [\textcolor{blue}{0.603}] & 0.930 [\textcolor{blue}{0.596}] & 0.927 [\textcolor{blue}{0.553}] & 0.984 [\textcolor{blue}{0.607}] \\
& HotpotQA & 0.794 & 0.623 & 0.977 [\textcolor{blue}{0.566}] & 0.987 [\textcolor{blue}{0.521}] & 0.936 [\textcolor{blue}{0.543}] & 0.966 [\textcolor{blue}{0.494}] & 0.993 [\textcolor{blue}{0.522}] \\
& NQ-Open  & 0.703 & 0.606 & 0.942 [\textcolor{blue}{0.671}] & 0.911 [\textcolor{blue}{0.544}] & 0.911 [\textcolor{blue}{0.601}] & 0.914 [\textcolor{blue}{0.536}] & 0.978 [\textcolor{blue}{0.575}] \\
& TriviaQA & 0.646 & 0.681 & 0.824 [\textcolor{blue}{0.817}] & 0.700 [\textcolor{blue}{0.690}] & 0.789 [\textcolor{blue}{0.760}] & 0.730 [\textcolor{blue}{0.701}] & 0.887 [\textcolor{blue}{0.882}] \\
\midrule
\multirow{4}{6em}{Mixtral-8x7B} 
& AmbigQA  & 0.760 & 0.666 & 0.948 [\textcolor{blue}{0.704}] & 0.891 [\textcolor{blue}{0.636}] & 0.955 [\textcolor{blue}{0.654}] & 0.944 [\textcolor{blue}{0.622}] & 0.975 [\textcolor{blue}{0.650}] \\
& HotpotQA & 0.761 & 0.657 & 0.970 [\textcolor{blue}{0.587}] & 0.966 [\textcolor{blue}{0.470}] & 0.930 [\textcolor{blue}{0.582}] & 0.970 [\textcolor{blue}{0.577}] & 0.987 [\textcolor{blue}{0.536}] \\
& NQ-Open  & 0.650 & 0.649 & 0.939 [\textcolor{blue}{0.652}] & 0.863 [\textcolor{blue}{0.517}] & 0.950 [\textcolor{blue}{0.590}] & 0.908 [\textcolor{blue}{0.529}] & 0.956 [\textcolor{blue}{0.563}] \\
& TriviaQA & 0.625 & 0.695 & 0.812 [\textcolor{blue}{0.800}] & 0.803 [\textcolor{blue}{0.754}] & 0.838 [\textcolor{blue}{0.818}] & 0.716 [\textcolor{blue}{0.725}] & 0.882 [\textcolor{blue}{0.858}] \\
\midrule
\multirow{4}{6em}{Mistral-7B} 
& AmbigQA  & 0.792 & 0.622 & 0.947 [\textcolor{blue}{0.730}] & 0.947 [\textcolor{blue}{0.627}] & 0.947 [\textcolor{blue}{0.628}] & 0.930 [\textcolor{blue}{0.523}] & 0.977 [\textcolor{blue}{0.647}] \\
& HotpotQA & 0.796 & 0.673 & 0.969 [\textcolor{blue}{0.649}] & 0.951 [\textcolor{blue}{0.478}] & 0.947 [\textcolor{blue}{0.680}] & 0.969 [\textcolor{blue}{0.578}] & 0.979 [\textcolor{blue}{0.673}] \\
& NQ-Open  & 0.726 & 0.639 & 0.925 [\textcolor{blue}{0.652}] & 0.919 [\textcolor{blue}{0.515}] & 0.939 [\textcolor{blue}{0.597}] & 0.920 [\textcolor{blue}{0.433}] & 0.976 [\textcolor{blue}{0.527}] \\
& TriviaQA & 0.718 & 0.608 & 0.879 [\textcolor{blue}{0.881}] & 0.863 [\textcolor{blue}{0.840}] & 0.822 [\textcolor{blue}{0.846}] & 0.735 [\textcolor{blue}{0.744}] & 0.927 [\textcolor{blue}{0.913}] \\
\bottomrule
\end{tabular}
\caption{Performance (Macro F1) of various evaluators across candidate LLMs and tasks. Values [in \textcolor{blue}{blue}] represent the reference-free mode.}
\label{tab:macro_f1_scores_free}
\end{table*}

\subsection{CLEV in multi-reference answers}
CLEV explicitly accommodates multiple gold reference answers by incorporating all available references into the judge LLM’s prompt during evaluation. For datasets like AmbigQA and TriviaQA, where questions often have multiple valid answers (e.g., synonyms, rephrased answers, or alternative factual representations), CLEV aggregates all reference answers into the judge’s input prompt (e.g., concatenating them as a comma-separated list). 

This design ensures that the judge evaluates the candidate’s output against the full spectrum of acceptable answers, mirroring the human evaluation protocol, where annotators are instructed to mark a response as correct if it aligns with any reference answer. However, as presented in our paper, LLM-based judges encounter challenges with multiple reference answers. This confusion is particularly evident in TriviaQA, where multiple reference answers introduce difficulties for the judges to recognize and evaluate a range of correct responses.

\subsection{Analysis of automatic metrics}
\begin{figure*}
    \centering
    \includegraphics[width=\textwidth]{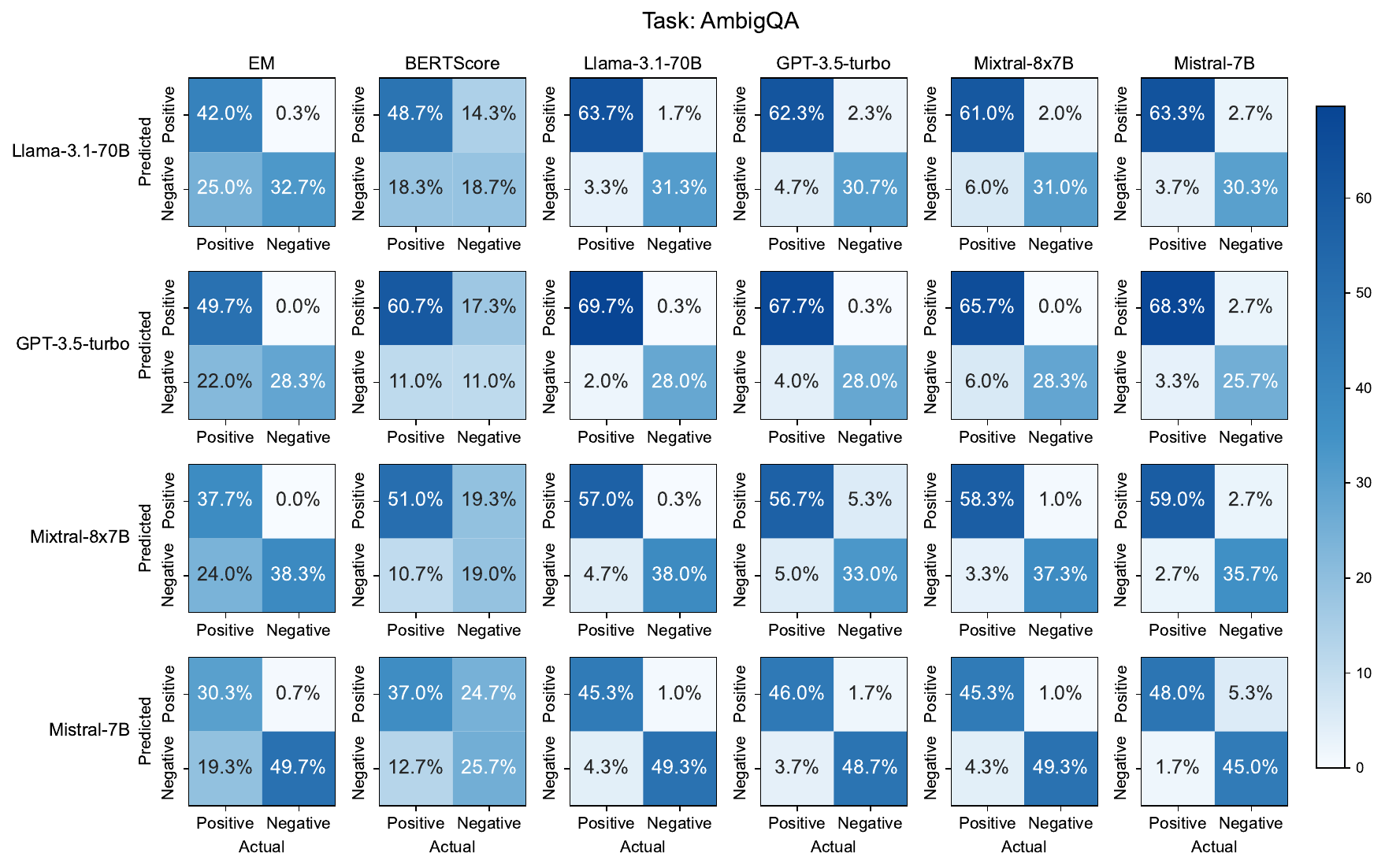}
    \caption{Confusion matrices comparing the performance of automatic metrics (EM, BERTScore) and individual LLM judges on AmbigQA.}
    \label{fig:ambigqa}
\end{figure*}

\begin{figure*}
    \centering
    \includegraphics[width=\textwidth]{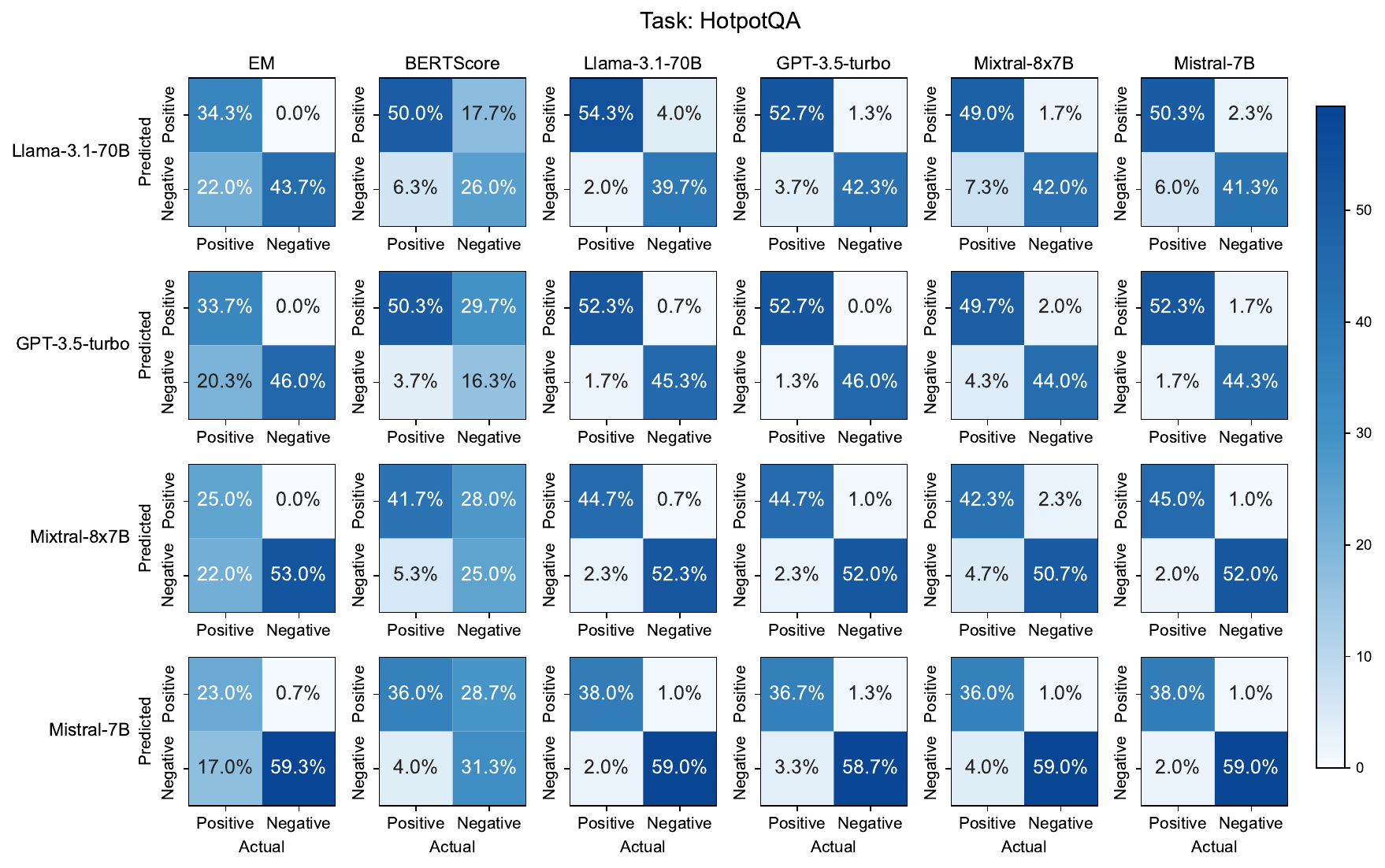}
    \caption{Confusion matrices comparing the performance of automatic metrics (EM, BERTScore) and individual LLM judges on HotpotQA.}
    \label{fig:hotpotqa}
\end{figure*}

\begin{figure*}
    \centering
    \includegraphics[width=\textwidth]{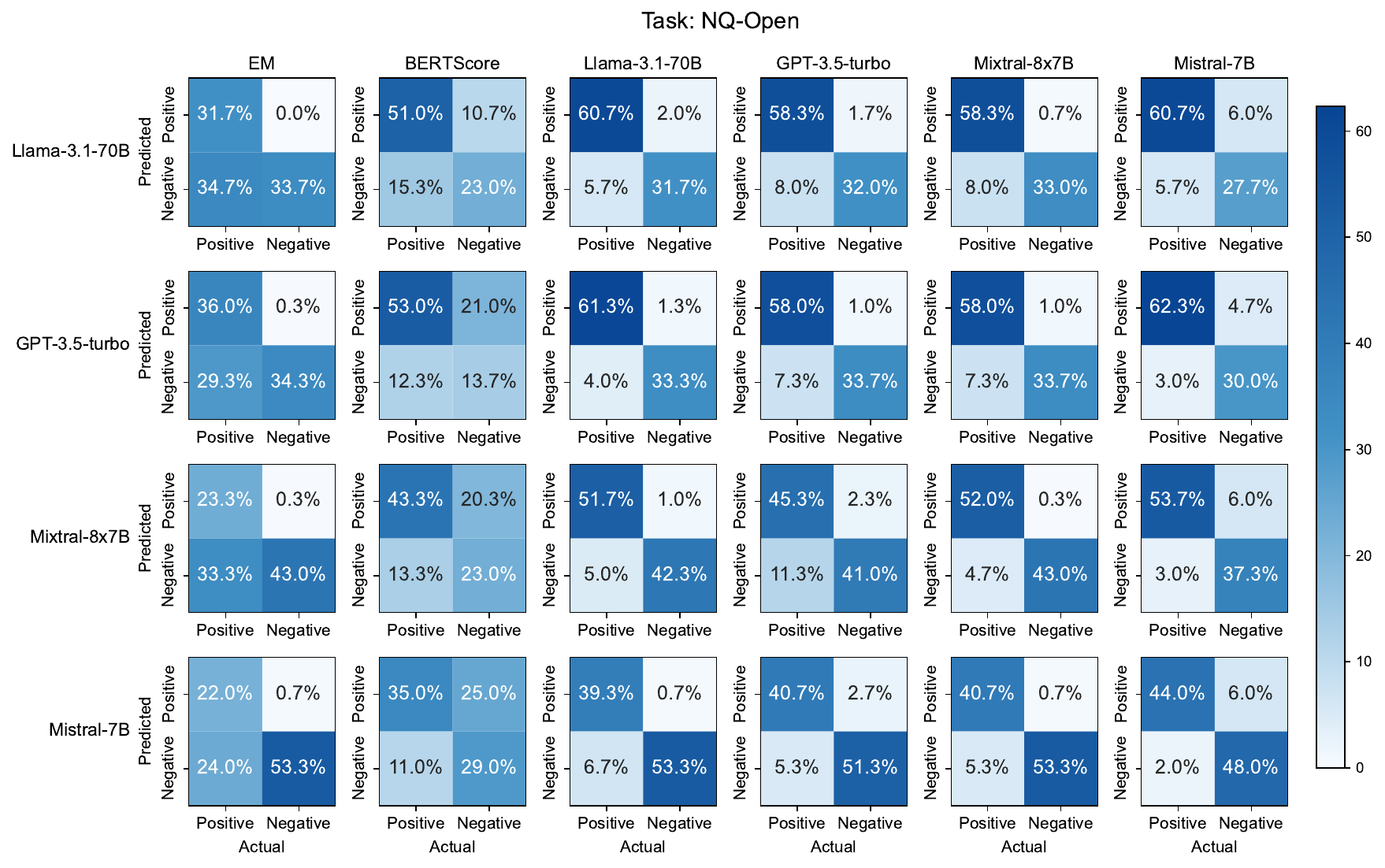}
    \caption{Confusion matrices comparing the performance of automatic metrics (EM, BERTScore) and individual LLM judges on NQ-Open.}
    \label{fig:nqopen}
\end{figure*}

\begin{figure*}
    \centering
    \includegraphics[width=\textwidth]{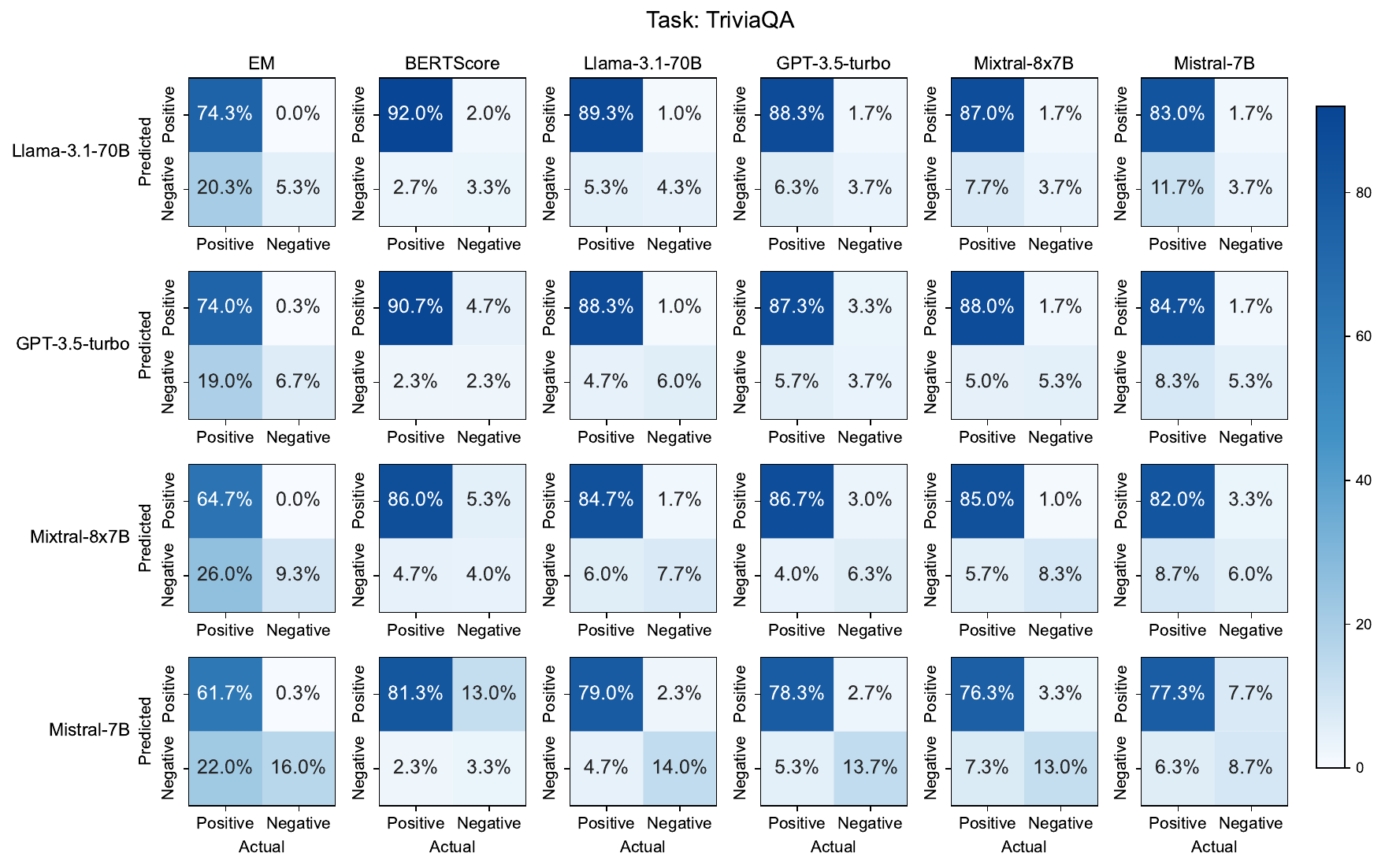}
    \caption{Confusion matrices comparing the performance of automatic metrics (EM, BERTScore) and individual LLM judges on TriviaQA.}
    \label{fig:triviaqa}
\end{figure*}

Figures~\ref{fig:ambigqa}, \ref{fig:hotpotqa}, \ref{fig:nqopen}, and \ref{fig:triviaqa} illustrate the fundamental trade-offs in automatic metrics. In TriviaQA, where multiple normalized reference answers exist, EM achieves impressive true positives (61.7-74.3\%) compared to HotpotQA (23.0-34.3\%) which contains single reference answers. EM's near-zero false positives across tasks (0-0.7\%) stem from its strict string matching – it only flags matches when answers are identical to references. Our error analysis found three primary causes of such rare false positives including preprocessing errors, where character normalization removes crucial distinctions, and reference ambiguities, where incomplete or ambiguous references lead to incorrect matches. Additionally, a semantic mismatch occurs when the EM incorrectly labels a prediction as true by matching text without considering its context. For instance, despite their different contextual meanings, EM wrongly marks a match between a model prediction of ``1944'' (describing the start of a war) and a reference answer containing ``1944'' (representing the end of the war).

EM string-matching guarantees high precision and makes EM particularly effective when exact wording is crucial, such as mathematical problems. However, its rigid criteria also result in substantial false negatives (17.0-34.7\%). These false negatives primarily occur when the candidate LLM generates semantically correct responses that differ from references in format or expression. Common cases include synonym usage and paraphrases, structural variations in phrasing (e.g., ``School of Medicine at Harvard'' vs. ``Harvard Medical School''), granularity discrepancies where answers differ in levels of detail from references (e.g., answering ``British writer'' instead of ``William Shakespeare''), and partial matches that contain valid information but don't exactly mirror the reference.

Unlike EM, BERTScore offers advantages in capturing semantic similarities. In TriviaQA, it gains high true positive rates (81.3-92.0\%) with relatively low false positives (2.0-13.0\%).  BERTScore's performance varies significantly across tasks and is influenced by its sensitivity to the threshold setting. In HotpotQA, where answers require multi-hop reasoning, true positives reach 36.0-50.3\%, with an increase in false positives (17.7-29.7\%). A similar pattern appears in NQ-Open, with true positives of 43.3-53.0\% and false positives of 10.7-21.0\%. Its tendency toward false positives indicates that relying solely on embedding similarity often accepts answers that are contextually related but factually incorrect. The false positives emerge through semantic drift (where similar embeddings yield false matches), contextual misalignment (where word meanings shift based on context), and threshold instability (where similarity cutoffs fail to distinguish subtle semantic differences). Additionally, false positives emerge due to the verbose responses where additional content artificially increases similarity scores.

\section{Prompting}\label{app:prompting}
In our main experiment, we performed zero-shot prompting in the two stages. First, we prompted candidate LLMs to obtain outputs for the given asks. Then, we instructed LLM judges to perform evaluations.  

\begin{figure*}[t]
\centering
\begin{tcolorbox}[colback=blue!5, colframe=black, title=\textbf{Candidate Prompt}, sharp corners=south, width=0.95\linewidth, boxrule=0.4pt]
\small
\textbf{You are a helpful assistant. Which comic book was also written by the writer of Crossed?}
\end{tcolorbox}
\caption{Prompt given to candidate GPT-3.5-turbo to elicit an answer for HotpotQA.}
\vspace{-5pt}
\label{fig:prompts_can_truth}
\end{figure*}

\begin{figure*}[t]
\centering
\begin{tcolorbox}[colback=blue!5, colframe=black, title=\textbf{Candidate Response}, sharp corners=south, width=0.95\linewidth, boxrule=0.4pt]
\small
Crossed was written by Garth Ennis. The Boys is a comic book series also written by Garth Ennis. So the comic book that was also written by the writer of Crossed is The Boys.
\end{tcolorbox}
\caption{Candidate GPT-3.5-turbo response to the prompt shown in Figure~\ref{fig:prompts_can_truth}.}
\vspace{-5pt}
\label{fig:prompts_can_truth_response}
\end{figure*}

\begin{figure*}[t]
\centering
\begin{tcolorbox}[colback=blue!5, colframe=black, title=\textbf{Judge Prompt}, sharp corners=south, width=0.95\linewidth, boxrule=0.4pt]
\small
You are a helpful assistant acting as an impartial judge. You will be given a Question and a Proposed Answer. Your task is to judge whether the Proposed Answer is correct by comparing it to the Reference Answer. If the Proposed Answer is correct, choose 'True', otherwise, choose 'False'. Provide a brief explanation for your decision. \\

Question: Which comic book was also written by the writer of Crossed? \\

Provided Answer: Crossed was written by Garth Ennis. The Boys is a comic book series also written by Garth Ennis. So the comic book that was also written by the writer of Crossed is The Boys. \\

Reference Answer: the Vertigo series ``Preacher'' \\

Evaluation: \\

Provide your response in the following format: \\
Decision: [True/False] \\
Explanation: [Your brief explanation] \\
\end{tcolorbox}
\caption{Prompting judge Llama-3.1-70B for evaluating the candidate GPT-3.5-turbo answer.}
\vspace{-5pt}
\label{fig:prompts_jud_truth}
\end{figure*}

\begin{figure*}[t]
\centering
\begin{tcolorbox}[colback=blue!5, colframe=black, title=\textbf{Judge Response}, sharp corners=south, width=0.95\linewidth, boxrule=0.4pt]
\small
\textbf{Decision:} False \\

Explanation: The Proposed Answer incorrectly identifies "The Boys" as the comic book written by the writer of Crossed. While it is true that Garth Ennis wrote Crossed, the Proposed Answer fails to consider other works written by Garth Ennis. The Reference Answer, "Preacher", is indeed another comic book series written by Garth Ennis, specifically for the Vertigo imprint.
\end{tcolorbox}
\caption{Llama-3.1-70B-Judge verdict on the candidate GPT-3.5-turbo answer.}
\vspace{-5pt}
\label{fig:prompts_jud_truth_response}
\end{figure*}

\end{document}